\documentclass[conference]{IEEEtran}
\IEEEoverridecommandlockouts

\newcommand\titletext{
Towards Formalizing HRI Data Collection Processes
}

\usepackage{graphicx}
\usepackage{xcolor}
\usepackage[numbers]{natbib}
\usepackage[hidelinks,pdftitle={\titletext}]{hyperref}
\usepackage{url}
\usepackage[textsize=tiny]{todonotes}
\usepackage{balance}
\usepackage{array}

\usepackage[utf8]{inputenc}
\usepackage{pifont}
\usepackage{newunicodechar}
\newunicodechar{✓}{\ding{51}}
\newunicodechar{✗}{\ding{55}}

\title{\titletext\thanks{This work has been supported in part by the Office of Naval Research under N00014-21-1-2418.}}

\author{
\IEEEauthorblockN{Zhao Han and Tom Williams}
\IEEEauthorblockA{\textit{MIRRORLab, Department of Computer Science, Colorado School of Mines}, Golden, Colorado, USA 80401\\
zhaohan@mines.edu, twilliams@mines.edu}}

\begin{document}

\maketitle

\begin{abstract}
Within the human-robot interaction (HRI) community, many researchers have focused on the careful design of human-subjects studies. However, other parts of the community, e.g., the technical advances community, also need to do human-subjects studies to \textit{collect data to train their models}, in ways that require user studies but without a strict experimental design. The design of such data collection is an underexplored area worthy of more attention. In this work, we contribute a clearly defined process to collect data with three steps for machine learning modeling purposes, grounded in recent literature, and detail an use of this process to facilitate the collection of a corpus of referring expressions. Specifically, we discuss our data collection goal and how we worked to encourage
well-covered and abundant participant responses, through our design of the task environment, the task itself, and the study procedure. We hope this work would lead to more data collection formalism efforts in the HRI community and a fruitful discussion during the workshop.
\end{abstract}

\section{Introduction}\label{sec:intro}

In a multidisciplinary field like HRI, it is important for researchers to leverage empirical research \cite{mackenzie2012human} to discover new knowledge from observations and experience. It is thus common to treat data collection solely within the lens of formal experimental design, to answer research questions by collecting, for example, qualitative data through interviews, surveys, or think-aloud protocols, and quantitative data from sensors or through coding qualitative data \cite{eatough2008interpretative,jost2020human,hoffman2020primer}.

Moreover, while data collected through user studies is increasingly made publicly available, such data is rarely reused. Instead, researchers in HRI tend to build on past datasets through new experiments to replicate that past work either tightly or with carefully controlled deviations, e.g., with other robots (\cite{bejerano2021back,ullman2021challenges}) or in different cultures (\cite{trovato2013cross,haring2014perception,andrist2015effects,strait2020three}). This paradigm has led to substantial recent research seeking to formalize experimental design~\cite{hoffman2020primer,bethel2020conducting} and analysis~\cite{bartlett2022have,schrum2020four} efforts within the unique contexts of HRI, with, unfortunately, data collection task design left behind.

Yet, other communities within HRI, such as the technical advances community, also collect human-subjects data, albeit for different purposes, such as collecting and modeling human data for more human-like and familiar interactions to improve robot experience \cite{siciliano2018humanoid,huggins2021practical}. 
For example, to advance social navigation, researchers have collected human navigation data to predict human activity \cite{rudenko2020human}, human-motion trajectory data (Th{\"o}r, \cite{rudenko2020thor}), and robot approaching behavior towards humans \cite{yang2020impact,nanavati2020autonomously}. Data in robots' view has also been collected to allow more practical robotics in unstructured environments \cite{martin2021jrdb,taylor2020robot}.

\begin{figure}
  \centering
  \includegraphics[width=0.8\columnwidth]{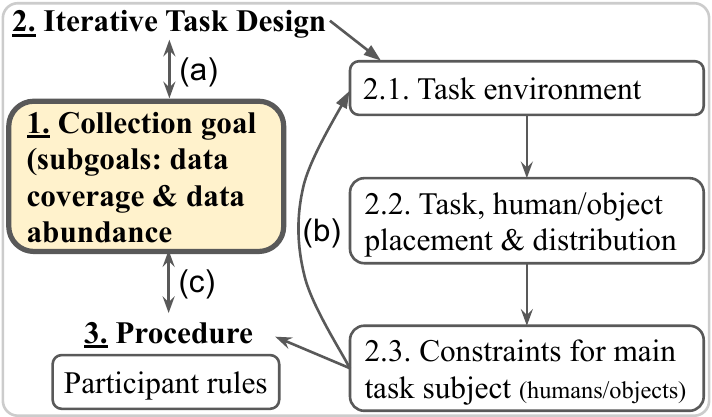}
  \caption{A flowchart for the process of designing a data collection study, grounded in recent HRI literature (See Table \ref{tab:t}). Collection goal and data coverage/abundance subgoals are first identified, followed by an iterative task design and procedure for collecting more data. Task design includes three major elements: the environment, the task with human/object configurations, and the criteria for the main task subject identified to reach the collection goal. The two bi-directional arrows, (a) and (c), and an iteration arrow (b) underline the importance of the iterative process}
  \label{fig:flowchart}
\end{figure}

In contrast to the experimental design and analysis works mentioned above, best practices or recommendations for collecting data for these types of purposes, i.e., machine learning modeling, have received less attention within HRI. In order to efficiently collect such data from human subjects, researchers must carefully design a task and a procedure to gather as many types of data as possible (i.e., data coverage) and solicit as much data as possible from participants (i.e., data abundance), as illustrated in Figure \ref{fig:flowchart} left. While the latter is key for data-hungry machine learning techniques, the former is vital for robustness to cover real-world scenarios.

In this workshop paper, we thus make two contributions:

\begin{enumerate}
\item A process for designing a data collection study grounded in recent HRI literature
\item A concrete study design example applied the process for a data collection in the context of human-robot dialogue
\end{enumerate}

This paper is organized as follows. We first give an overview of the process and its three steps in Section \ref{sec:process} with example works. We then detail the case study in Section \ref{sec:casestudy}, whose task design is more specifically discussed in other recent work~\cite{han2022task}. Following the workflow shown in Figure \ref{fig:flowchart}, Section \ref{sec:casestudy} first describes the goal that guides us and defines data coverage in our domain. We then discuss how we designed the task to reach our goal, including task environment and task choice, and object placement and distribution. A brief discussion on the iterative process is also provided. Lastly, we conclude with insights as to how we used extra rules in our procedure to encourage more data to be collected from participants.

\section{A Process for Data Collection Study Design}\label{sec:process}

As shown in Figure \ref{fig:flowchart}, the process consists of three steps: goal, task design, and procedure.
First, the \textbf{goal} of the data collection effort needs to be identified. Akin to hypotheses that guide the design of human-subject experiments, it is critical to articulate the precise goals that guide non-experimental data collection efforts. The goal is dependent on the domain or application of the proposed machine learning model. For example, in \citet{taylor2020robot}'s work, the goal was to collect egocentric color and depth (RGB-D) data of groups of people to predict social groups. In \citet{yang2020impact}'s work, the goal was to collect different reactions from humans when a robot approaches from different directions to join a conversation.

Concretely, the goal can be divided into two subgoals: data coverage and data abundance. \textbf{Data coverage} concerns different types of or forms of data that should be collected. For example, in \citet{taylor2020robot}'s work, the data was recorded in multiple crowded, sunny, outdoor environments, covering occlusion, shadow, lighting, and motion patterns to handle real-world challenges. In \citet{yang2020impact}'s work, the authors collected data from two group types, nine approaching directions, and three Wizard-of-Oz robot styles.

While data coverage addresses data quality, \textbf{data abundance} addresses data quantity. While \citet{taylor2020robot} do not explicitly discuss this, they collected 1.5 hours of 16,827 RGB-D frames. In \citet{yang2020impact}'s work on modeling conversational approaching behavior, the authors used 16 on-body cameras and a Motion Capture suit with 37 markers to gather more data from participants. As we show in our case study, data abundance can also be achieved by deliberately soliciting more data from participants.

Secondly, a \textbf{task design} specifically for data collection must be carefully constructed to reach the goal. The task design includes the environment, the task itself with human or object placement and distribution, and some criteria for the main task subject. Because \citet{taylor2020robot}'s work studies crowd behavior in a public environment, this step was skipped. In \citet{yang2020impact}'s work, the authors use a three-person ``Who's the Spy'' game with the robot being adjudicator to identify the spy. While without physical objects, the \textbf{task environment} consists of a marked circle that a triad of participants stands on; The robot stands at room corners outside of the circle. The \textbf{task} for each participant was to describe the material of the word on a card given to them. While objects are not the focus, standing participants face each other and were distributed in the center of a room. The \textbf{main task subject} is the robot that was constrained to only be teleoperated to approach in different directions to join the group when the spy is identified. In our case study, our main task subjects were buildings and we explicitly imposed additional constraints.

Lastly, a well-thought procedure needs to be in place to reach the collection goal. In \citet{yang2020impact}'s work, participants were asked to stand at fixed positions so they are in the field of view of the cameras. As \citet{taylor2020robot} studies public groups of people, no explicit procedure was given.

Figure \ref{fig:flowchart}'s bi-directional arrows and cyclic nature emphasizes the iterative nature of the process centered around a primary collection goal. Yet this iterative process is not typically reported in the literature, similar to user studies where pilot studies may not be reported. In the following section, we will detail our case study and detail the iterative process we followed.

\newcolumntype{P}[1]{>{\hspace{0pt}}p{#1}}
\begin{table}
\caption{Sample works fitting into the proposed process (Task*: environment, task, and main subject constraints)}
\label{tab:t}
\begin{center}
\scriptsize
\begin{tabular}{lllllll}
\hline
& \textbf{Domain} & \textbf{Goal} & \textbf{Cover.} & \textbf{Abundance} & \textbf{Task*} & \textbf{Procedure}\\
\hline
\cite{taylor2020robot}	    &Group&✓&✓&✗&NA&NA\\
\hline
\cite{yang2020impact}  &Navigation&✓&✓&✓ &✓✓✓&✓\\
\hline
\cite{nanavati2020autonomously} &Service&✓&✓&✗&✓✓✓&✓\\
\hline
\cite{rudenko2020thor}  &Navigation&✓&✓&✓&✓✓✓&✓\\
\hline
\cite{chen2020trust} &Trust&✓&✓&✗&✓✓✗&✓\\
\hline
\cite{martin2021jrdb}  &Perception&✓&✓&✓&NA&NA\\
\hline
\cite{engwall2021identification} &Tutoring&✓&✓&✗&✓✓✓&✓\\
\hline
\cite{novoa2021automatic} &Speech&✓&✓&✗&✓✓✓&✓\\
\hline
\end{tabular}
\end{center}
\end{table}

In addition to the two examples, Table \ref{tab:t} lists six more HRI conference, HRI workshop, and ACM THRI papers from 2018--2021 (filtered with the ``data collection'' keyword), and whether they fit the proposed process. It is worth noting that only eight papers were found, indicating the need for this work. Moreover, over half of them did not discuss data abundance.

\section{A Case Study on Applying the Process}\label{sec:casestudy}

With the process defined with concise, grounded examples, we can now detail a case study from our recent work \cite{han2022task} that applied the described process in the human-robot dialog domain.

\subsection{Collection Goal and Data Coverage}\label{sec:goal}

The goal in our example was to collect a corpus of verbal and nonverbal data that is rich and varied yet representative of typical human dialogue patterns. Second, we aimed to collect referring forms and gestures picking out both present, perceivable entities, as well as entities that are \textit{not in the scene} or \textit{were in the scene but are no longer visible} as humans are moved away from the scene.
Third, we aimed to collect data that would cover a wide range of referring forms, including pronominal, deictic, and definite forms (e.g., it, this, that, this-N', that-N', and the-N'), as well as indefinite forms such as a-N'. Finally, we wished to collect data that was rich in nonverbal cues like gestures. The data abundance will be described in Section \ref{sec:procedure} of the procedure.

\subsection{Iterative Task Design}

\begin{figure}
  \centering
  \includegraphics[width=0.9\columnwidth]{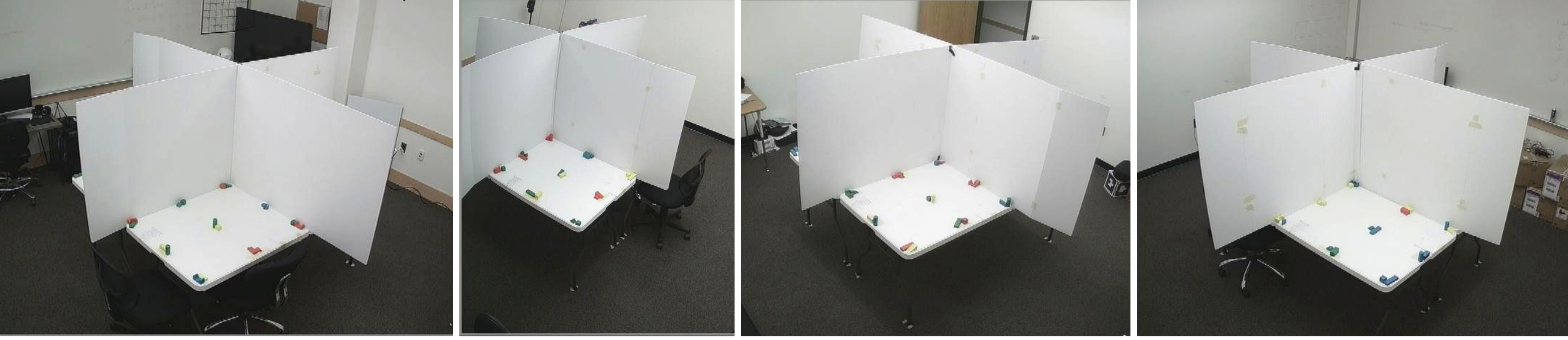}
  \caption{The four-quadrant task environment, adapted from the $2 \times 2$ video multiplexer from four cameras installed at the corners of the room's ceiling. The environment is made by adjoining two tables and four foam boards, which are longer than the table to avoid participant looking into other quadrants.}
  \label{fig:taskenv:birdview}
\end{figure}

\textit{Task Environment}: Different from the single tabletop scenarios in previous robot dialogue research \cite{roy2005semiotic,hsiao2008object,matuszek2014learning,scalise2018natural} where all objects are present in a robot's operating environment, we used a four-quadrant tabletop scenario (Figure \ref{fig:taskenv:birdview}) by adjoining two tables \cite{walmarttable} and separating them into four quadrants with two long foam boards \cite{targetfoamboard}, so objects can be hidden in different quadrants and referers can refer to both present and non-present objects.

\begin{figure}
  \centering
  \includegraphics[width=0.8\columnwidth]{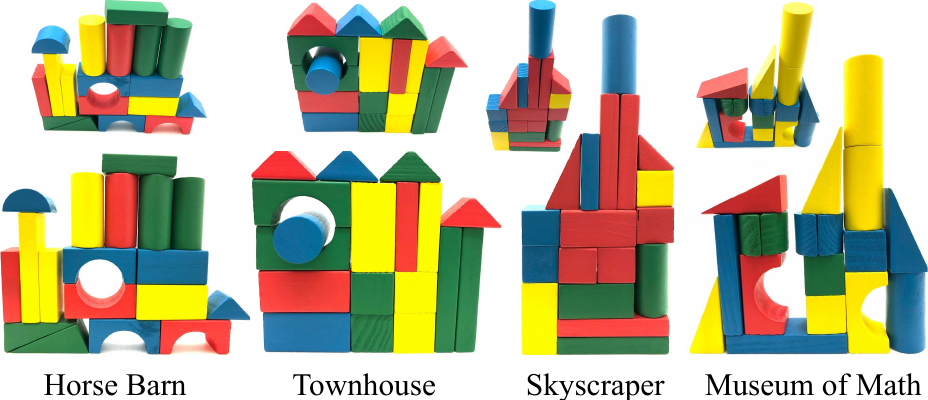}
  \caption{Four buildings to be constructed. To help participants identify individual blocks, two angles were provided. Each building has repeated blocks to reach the data collection goal, i.e., wider variety of referring forms.}
  \label{fig:buildings}
\end{figure}

\textit{Task}: The task environment helps reach the goals of encouraging references to both visible and non-visible objects; similarly, the task, number of objects, and object distribution in quadrants should help us to collect more natural language references and gestures.
To that end, we chose a series of collaborative \textit{tower building} tasks \cite{jung2020robot} where instructor participants teach learner participants construct four buildings (Figure \ref{fig:buildings}) from $18 \times 4 = 72$ blocks \cite{blockproduct} in different quadrants. The repetitive elements during the building process increase use of reference forms, either in speech or with gestures.

\textit{Object Placement and Distribution}\label{sec:block}: We used a number of block shapes, including triangles, cubes, cuboids, cylinders, arches, and half-circles, so they are not too complex to describe and participants can focus on referring to them in the same quadrant or previous quadrants.
The blocks required to construct each building are randomly placed at the vertices of a $3 \times 3$ grid. This placement strategy leads to varying the physical distance between blocks and encourages referring to visible objects with ``this'' and ``that'' \cite{dixon2003demonstratives}.

\textit{Criteria for Main Task Subject}: To cover indefinite nouns (e.g., a N), we constrained the placement of the blocks used to construct buildings as follows: Half of the blocks needed for each building are distributed to the quadrant in which that building is to be constructed, and the other half of the blocks need to be evenly distributed in the other three quadrants.
To meet this constraint, each building has an even number of 18 blocks. \textit{Nine of them} are placed in the quadrant where the building is constructed, and each of the other three quadrants has 3 (i.e., $\frac{9}{3}$) blocks, depleting the remaining nine blocks.

\begin{figure}
  \centering
  \includegraphics[width=0.5\columnwidth]{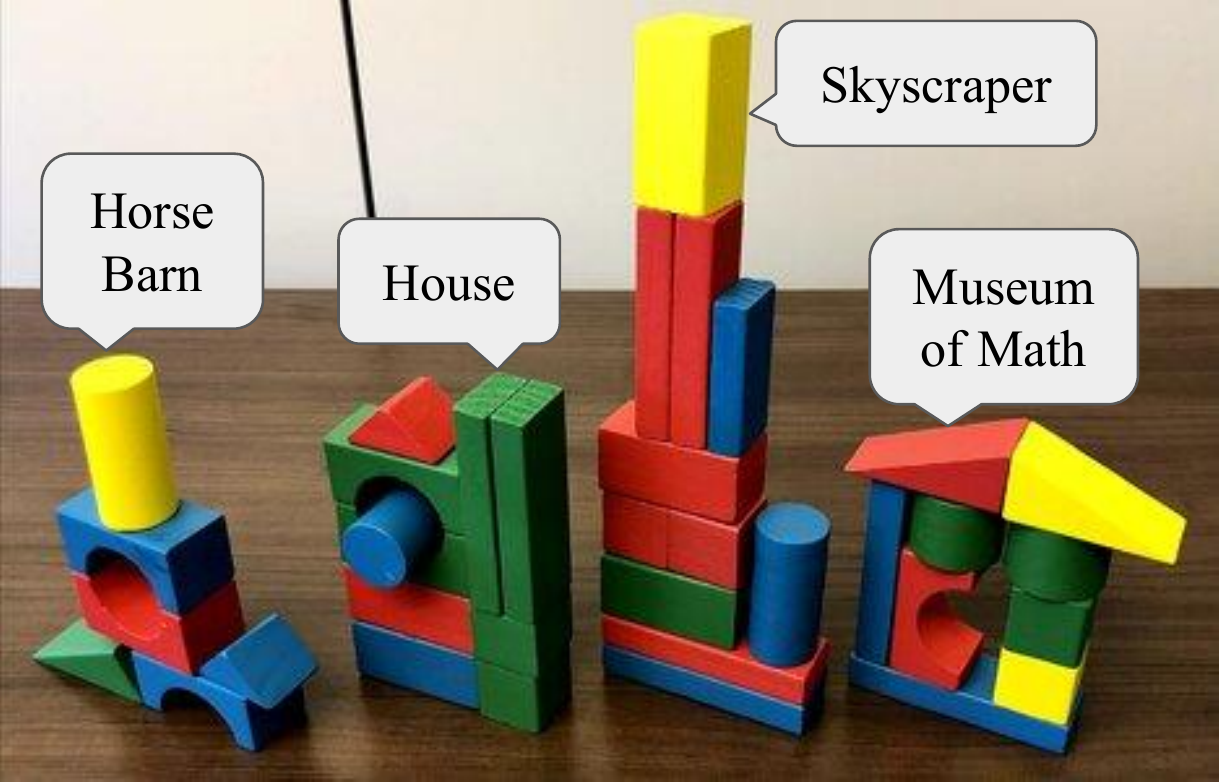}
  \caption{The \underline{\textbf{first iteration}} of simpler buildings. The task design is an iterative process by gathering feedback and incorporating improvements towards the goal of our collection effort.}
  \label{fig:initialbuildings}
\end{figure}

\textit{Iterative Task Design Process}: The task design is an iterative process, similar to interaction design \cite{rodgers2011interaction}. Feedback and improvement should be incorporated before the design is finalized. Indeed, the buildings were previously much simpler and consisted of fewer blocks, as shown in Figure \ref{fig:initialbuildings}, but later made more complex to encourage the production of more references and gestures by participants in a single session. To gather feedback, we ran pilot studies and presented the task design in lab meetings.

\subsection{Procedure Design}\label{sec:procedure}

With the task design in place, we can describe how we designed our study procedure.
Once seated, both participants were provided with rule cards as reminders. However, only instructors' cards included the building photos, not visible to the learner participants, encouraging more speech and gestures from instructor participants. Moreover, learners were asked not to speak unless absolutely necessary to proceed, limiting data provided on their part but significantly increasing the amount of language needed to be used by instructors. Similarly, instructors were asked not to touch any blocks, and to only ask learners to find blocks if those blocks were not found in the current quadrant. These tactics encouraged additional language and gestures by instructors, and encouraged instructors to visually search their quadrants before issuing instructions, so as to encourage a wider variety of referring forms.

\section{Discussion and Conclusions}

As we have mentioned, our experimental design process was iterative, and was not perfect in the beginning. The flowchart we provide is not a precise recipe that must be followed exactly (as seen that some examples did not follow part of the process). But instead it is provided to make the logic of the design process more clear, serving as a clearer takeaway from this work. Indeed, the experiment design for data collection requires creativity, especially for the task. Hopefully, our work will inspire HRI researchers to step outside of the boundary of the well-established user studies and work more on data collection to develop human-like and familiar interactions.

In conclusion, we contribute a formalized process for a model data collection experiment, informed by recent HRI literature. Centered around reaching a high-level data collection goal, as well as sub-goals regarding data coverage, variety, and abundance, we followed the familiar task design and procedure elements used in traditional experimental designs. We provided a detailed account of the underlying design considerations, and a flow chart that visualizes the steps.
In the future, we would like to expand this workshop paper to a comprehensive meta-analysis of data collection work in HRI and a taxonomy for the the task and procedure design.

\bibliographystyle{IEEEtranN}
\balance
\bibliography{bib}

\begin{thebibliography}{35}
\providecommand{\natexlab}[1]{#1}
\providecommand{\url}[1]{#1}
\csname url@samestyle\endcsname
\providecommand{\newblock}{\relax}
\providecommand{\bibinfo}[2]{#2}
\providecommand{\BIBentrySTDinterwordspacing}{\spaceskip=0pt\relax}
\providecommand{\BIBentryALTinterwordstretchfactor}{4}
\providecommand{\BIBentryALTinterwordspacing}{\spaceskip=\fontdimen2\font plus
\BIBentryALTinterwordstretchfactor\fontdimen3\font minus
  \fontdimen4\font\relax}
\providecommand{\BIBforeignlanguage}[2]{{%
\expandafter\ifx\csname l@#1\endcsname\relax
\typeout{** WARNING: IEEEtranN.bst: No hyphenation pattern has been}%
\typeout{** loaded for the language `#1'. Using the pattern for}%
\typeout{** the default language instead.}%
\else
\language=\csname l@#1\endcsname
\fi
#2}}
\providecommand{\BIBdecl}{\relax}
\BIBdecl

\bibitem[MacKenzie(2012)]{mackenzie2012human}
I.~S. MacKenzie, \emph{Human-computer interaction: An empirical research
  perspective}.\hskip 1em plus 0.5em minus 0.4em\relax Newnes, 2012.

\bibitem[Eatough and Smith(2008)]{eatough2008interpretative}
V.~Eatough and J.~A. Smith, ``Interpretative phenomenological analysis,''
  \emph{The Sage handbook of qualitative research in psychology}, vol. 179, p.
  194, 2008.

\bibitem[Jost et~al.(2020)Jost, Le~P{\'e}v{\'e}dic, Belpaeme, Bethel,
  Chrysostomou, Crook, Grandgeorge, and Mirnig]{jost2020human}
C.~Jost, B.~Le~P{\'e}v{\'e}dic, T.~Belpaeme, C.~Bethel, D.~Chrysostomou,
  N.~Crook, M.~Grandgeorge, and N.~Mirnig, \emph{Human-Robot Interaction:
  Evaluation Methods and Their Standardization}.\hskip 1em plus 0.5em minus
  0.4em\relax Springer Nature, 2020, vol.~12.

\bibitem[Hoffman and Zhao(2020)]{hoffman2020primer}
G.~Hoffman and X.~Zhao, ``A primer for conducting experiments in human--robot
  interaction,'' \emph{ACM Transactions on Human-Robot Interaction (THRI)},
  vol.~10, no.~1, pp. 1--31, 2020.

\bibitem[Bejerano et~al.(2021)Bejerano, Robinette, Yanco, and
  Phillips]{bejerano2021back}
G.~Bejerano, P.~Robinette, H.~A. Yanco, and E.~Phillips, ``Back to the future:
  Opinions of autonomous cars over time,'' in \emph{Companion of the 2021
  ACM/IEEE International Conference on Human-Robot Interaction}, 2021, pp.
  157--161.

\bibitem[Ullman et~al.(2021)Ullman, Aladia, and Malle]{ullman2021challenges}
D.~Ullman, S.~Aladia, and B.~F. Malle, ``Challenges and opportunities for
  replication science in hri: A case study in human-robot trust,'' in
  \emph{Proceedings of the 2021 ACM/IEEE International Conference on
  Human-Robot Interaction}, 2021, pp. 110--118.

\bibitem[Trovato et~al.(2013)Trovato, Zecca, Sessa, Jamone, Ham, Hashimoto, and
  Takanishi]{trovato2013cross}
G.~Trovato, M.~Zecca, S.~Sessa, L.~Jamone, J.~Ham, K.~Hashimoto, and
  A.~Takanishi, ``Cross-cultural study on human-robot greeting interaction:
  acceptance and discomfort by egyptians and japanese,'' \emph{Paladyn, Journal
  of Behavioral Robotics}, vol.~4, no.~2, pp. 83--93, 2013.

\bibitem[Haring et~al.(2014)Haring, Silvera-Tawil, Matsumoto, Velonaki, and
  Watanabe]{haring2014perception}
K.~S. Haring, D.~Silvera-Tawil, Y.~Matsumoto, M.~Velonaki, and K.~Watanabe,
  ``Perception of an android robot in japan and australia: A cross-cultural
  comparison,'' in \emph{International conference on social robotics}.\hskip
  1em plus 0.5em minus 0.4em\relax Springer, 2014, pp. 166--175.

\bibitem[Andrist et~al.(2015)Andrist, Ziadee, Boukaram, Mutlu, and
  Sakr]{andrist2015effects}
S.~Andrist, M.~Ziadee, H.~Boukaram, B.~Mutlu, and M.~Sakr, ``Effects of culture
  on the credibility of robot speech: A comparison between english and
  arabic,'' in \emph{Proceedings of the Tenth Annual ACM/IEEE International
  Conference on Human-Robot Interaction}, 2015, pp. 157--164.

\bibitem[Strait et~al.(2020)Strait, Lier, Bernotat, Wachsmuth, Eyssel,
  Goldstone, and {\v{S}}abanovi{\'c}]{strait2020three}
M.~Strait, F.~Lier, J.~Bernotat, S.~Wachsmuth, F.~Eyssel, R.~Goldstone, and
  S.~{\v{S}}abanovi{\'c}, ``A three-site reproduction of the joint simon effect
  with the nao robot,'' in \emph{Proceedings of the 2020 ACM/IEEE International
  Conference on Human-Robot Interaction}, 2020, pp. 103--111.

\bibitem[Bethel et~al.(2020)Bethel, Henkel, and Baugus]{bethel2020conducting}
C.~L. Bethel, Z.~Henkel, and K.~Baugus, ``Conducting studies in human-robot
  interaction,'' in \emph{Human-Robot Interaction}.\hskip 1em plus 0.5em minus
  0.4em\relax Springer, 2020, pp. 91--124.

\bibitem[Bartlett et~al.(2022)Bartlett, Edmunds, Belpaeme, and
  Thill]{bartlett2022have}
M.~E. Bartlett, C.~Edmunds, T.~Belpaeme, and S.~Thill, ``Have i got the power?
  analysing and reporting statistical power in hri,'' \emph{ACM Transactions on
  Human-Robot Interaction (THRI)}, vol.~11, no.~2, pp. 1--16, 2022.

\bibitem[Schrum et~al.(2020)Schrum, Johnson, Ghuy, and
  Gombolay]{schrum2020four}
M.~L. Schrum, M.~Johnson, M.~Ghuy, and M.~C. Gombolay, ``Four years in review:
  Statistical practices of likert scales in human-robot interaction studies,''
  in \emph{Companion of the 2020 ACM/IEEE International Conference on
  Human-Robot Interaction}, 2020, pp. 43--52.

\bibitem[Siciliano and Khatib(2018)]{siciliano2018humanoid}
B.~Siciliano and O.~Khatib, ``Humanoid robots: historical perspective, overview
  and scope,'' \emph{Humanoid robotics: a reference}, pp. 1--6, 2018.

\bibitem[Huggins et~al.(2021)Huggins, Alghowinem, Jeong, Colon-Hernandez,
  Breazeal, and Park]{huggins2021practical}
M.~Huggins, S.~Alghowinem, S.~Jeong, P.~Colon-Hernandez, C.~Breazeal, and H.~W.
  Park, ``Practical guidelines for intent recognition: Bert with minimal
  training data evaluated in real-world hri application,'' in \emph{Proceedings
  of the 2021 ACM/IEEE International Conference on Human-Robot Interaction},
  2021, pp. 341--350.

\bibitem[Rudenko et~al.(2020{\natexlab{a}})Rudenko, Palmieri, Herman, Kitani,
  Gavrila, and Arras]{rudenko2020human}
A.~Rudenko, L.~Palmieri, M.~Herman, K.~M. Kitani, D.~M. Gavrila, and K.~O.
  Arras, ``Human motion trajectory prediction: A survey,'' \emph{The
  International Journal of Robotics Research}, vol.~39, no.~8, pp. 895--935,
  2020.

\bibitem[Rudenko et~al.(2020{\natexlab{b}})Rudenko, Kucner, Swaminathan,
  Chadalavada, Arras, and Lilienthal]{rudenko2020thor}
A.~Rudenko, T.~P. Kucner, C.~S. Swaminathan, R.~T. Chadalavada, K.~O. Arras,
  and A.~J. Lilienthal, ``Th{\"o}r: human-robot navigation data collection and
  accurate motion trajectories dataset,'' \emph{IEEE Robotics and Automation
  Letters}, vol.~5, no.~2, pp. 676--682, 2020.

\bibitem[Yang et~al.(2020)Yang, Yin, Bj{\"o}rkman, and Peters]{yang2020impact}
F.~Yang, W.~Yin, M.~Bj{\"o}rkman, and C.~Peters, ``Impact of trajectory
  generation methods on viewer perception of robot approaching group
  behaviors,'' in \emph{2020 29th IEEE International Conference on Robot and
  Human Interactive Communication (RO-MAN)}.\hskip 1em plus 0.5em minus
  0.4em\relax IEEE, 2020, pp. 509--516.

\bibitem[Nanavati et~al.(2020)Nanavati, Doering, Br{\v{s}}{\v{c}}i{\'c}, and
  Kanda]{nanavati2020autonomously}
A.~Nanavati, M.~Doering, D.~Br{\v{s}}{\v{c}}i{\'c}, and T.~Kanda,
  ``Autonomously learning one-to-many social interaction logic from human-human
  interaction data,'' in \emph{Proceedings of the 2020 ACM/IEEE International
  Conference on Human-Robot Interaction}, 2020, pp. 419--427.

\bibitem[Mart{\'\i}n-Mart{\'\i}n et~al.(2021)Mart{\'\i}n-Mart{\'\i}n, Patel,
  Rezatofighi, Shenoi, Gwak, Frankel, Sadeghian, and Savarese]{martin2021jrdb}
R.~Mart{\'\i}n-Mart{\'\i}n, M.~Patel, H.~Rezatofighi, A.~Shenoi, J.~Gwak,
  E.~Frankel, A.~Sadeghian, and S.~Savarese, ``{JRDB}: A dataset and benchmark
  of egocentric robot visual perception of humans in built environments,''
  \emph{IEEE Transactions on Pattern Analysis and Machine Intelligence}, 2021.

\bibitem[Taylor et~al.(2020)Taylor, Chan, and Riek]{taylor2020robot}
A.~Taylor, D.~M. Chan, and L.~D. Riek, ``Robot-centric perception of human
  groups,'' \emph{ACM Transactions on Human-Robot Interaction (THRI)}, vol.~9,
  no.~3, pp. 1--21, 2020.

\bibitem[Han and Williams(2022)]{han2022task}
Z.~Han and T.~Williams, ``A task design for studying referring behaviors for
  linguistic {HRI},'' in \emph{Companion of the 2022 ACM/IEEE International
  Conference on Human-Robot Interaction}.\hskip 1em plus 0.5em minus
  0.4em\relax IEEE, 2022.

\bibitem[Chen et~al.(2020)Chen, Nikolaidis, Soh, Hsu, and
  Srinivasa]{chen2020trust}
M.~Chen, S.~Nikolaidis, H.~Soh, D.~Hsu, and S.~Srinivasa, ``Trust-aware
  decision making for human-robot collaboration: Model learning and planning,''
  \emph{ACM Transactions on Human-Robot Interaction (THRI)}, vol.~9, no.~2, pp.
  1--23, 2020.

\bibitem[Engwall et~al.(2021)Engwall, Cumbal, {\'A}guas~Lopes, Ljung, and
  M{\aa}nsson]{engwall2021identification}
O.~Engwall, R.~Cumbal, J.~D. {\'A}guas~Lopes, M.~Ljung, and L.~M{\aa}nsson,
  ``Identification of low-engaged learners in robot-led second language
  conversations with adults,'' \emph{ACM Transactions on Human-Robot
  Interaction}, 2021.

\bibitem[Novoa et~al.(2021)Novoa, Mahu, Wuth, Escudero, Fredes, and
  Yoma]{novoa2021automatic}
J.~Novoa, R.~Mahu, J.~Wuth, J.~P. Escudero, J.~Fredes, and N.~B. Yoma,
  ``Automatic speech recognition for indoor hri scenarios,'' \emph{ACM
  Transactions on Human-Robot Interaction (THRI)}, vol.~10, no.~2, pp. 1--30,
  2021.

\bibitem[Roy(2005)]{roy2005semiotic}
D.~Roy, ``Semiotic schemas: A framework for grounding language in action and
  perception,'' \emph{Artificial Intelligence}, vol. 167, no. 1-2, pp.
  170--205, 2005.

\bibitem[Hsiao et~al.(2008)Hsiao, Vosoughi, Tellex, Kubat, and
  Roy]{hsiao2008object}
K.-y. Hsiao, S.~Vosoughi, S.~Tellex, R.~Kubat, and D.~Roy, ``Object schemas for
  responsive robotic language use,'' in \emph{Proceedings of the 3rd ACM/IEEE
  international conference on Human robot interaction}, 2008, pp. 233--240.

\bibitem[Matuszek et~al.(2014)Matuszek, Bo, Zettlemoyer, and
  Fox]{matuszek2014learning}
C.~Matuszek, L.~Bo, L.~Zettlemoyer, and D.~Fox, ``Learning from unscripted
  deictic gesture and language for human-robot interactions,'' in
  \emph{Twenty-Eighth AAAI Conference on Artificial Intelligence}, 2014.

\bibitem[Scalise et~al.(2018)Scalise, Li, Admoni, Rosenthal, and
  Srinivasa]{scalise2018natural}
R.~Scalise, S.~Li, H.~Admoni, S.~Rosenthal, and S.~S. Srinivasa, ``Natural
  language instructions for human--robot collaborative manipulation,''
  \emph{The International Journal of Robotics Research}, vol.~37, no.~6, pp.
  558--565, 2018.

\bibitem[wal()]{walmarttable}
``Mainstays 6 foot fold-in-half table, white granite,''
  \url{https://www.walmart.com/ip/622822527}, accessed: 2022-02-20.

\bibitem[tar()]{targetfoamboard}
``Elmer's 36" x 48" tri-fold foam presentation board - white,''
  \url{https://www.target.com/p/A-13313406}, accessed: 2022-02-20.

\bibitem[Jung et~al.(2020)Jung, DiFranzo, Shen, Stoll, Claure, and
  Lawrence]{jung2020robot}
M.~F. Jung, D.~DiFranzo, S.~Shen, B.~Stoll, H.~Claure, and A.~Lawrence,
  ``Robot-assisted tower construction—a method to study the impact of a
  robot’s allocation behavior on interpersonal dynamics and collaboration in
  groups,'' \emph{ACM Transactions on Human-Robot Interaction (THRI)}, vol.~10,
  no.~1, pp. 1--23, 2020.

\bibitem[blo()]{blockproduct}
``100 piece wood blocks set,''
  \url{https://www.melissaanddoug.com/100-piece-wood-blocks-set/481.html},
  accessed: 2022-02-20.

\bibitem[Dixon(2003)]{dixon2003demonstratives}
R.~M. Dixon, ``Demonstratives: A cross-linguistic typology,'' \emph{Studies in
  Language. International Journal sponsored by the Foundation “Foundations of
  Language”}, vol.~27, no.~1, pp. 61--112, 2003.

\bibitem[Rodgers et~al.(2011)Rodgers, Sharp, and
  Preece]{rodgers2011interaction}
Y.~Rodgers, H.~Sharp, and J.~Preece, ``Interaction design: Beyond
  human-computer interaction,'' 2011.

\end{thebibliography}

\clearpage
\end{document}